\pdfoutput=1
\documentclass{article}

\usepackage{aaai18}
\usepackage{times}
\usepackage{multirow}
\usepackage{hyperref}
\usepackage{graphicx}
\usepackage{subfigure}
\usepackage{amsmath}

\title{Network of Recurrent Neural Networks}

\author{Chao-Ming Wang\\ 
School of Software, Beijing Jiaotong University, Beijing, China  \\
\href{mailto:oujago@gmail.com}{oujago@gmail.com}}

\begin{document}

\maketitle

\begin{abstract}
We describe a class of systems theory based neural networks called ``Network Of Recurrent neural networks'' (NOR), which introduces a new structure level to RNN related models. In NOR, RNNs are viewed as the high-level neurons and are used to build the high-level layers. More specifically, we propose several methodologies to design different NOR topologies according to the theory of system evolution. Then we carry experiments on three different tasks to evaluate our implementations. Experimental results show our models outperform simple RNN remarkably under the same number of parameters, and sometimes achieve even better results than GRU and LSTM.
\end{abstract}

\section{Introduction}

In recent years, Recurrent Neural Networks (RNNs) \cite{elman1990finding} have been widely used in Natural Language Processing (NLP). Traditionally, RNNs are directly used to build the final models. In this paper, we propose a novel idea called ``Network Of Recurrent neural networks'' (NOR) which utilizes existing basic RNN layers to make the structure design of the higher-level layers. From a standpoint of systems theory \cite{von1968general,von1972history}, a recurrent neural network is a group or an organization made up of a number of interacting parts, and it actually is viewed as a \textit{complex system}, or a \textit{complexity}. Dialectically, every system is relative: it is not only the system of its parts, but also the part of a larger system. In NOR structures, RNN is viewed as the high-level neuron, and several high-level neurons are used to build the high-level layers rather than directly used to construct the whole models. 
 
Conventionally, there are three levels of structure in Deep Neural Networks (DNNs): \textit{neurons}, \textit{layers} and \textit{whole nets} (or called \textit{models}). From a perspective of systems theory, at each level of such increasing complexity, novel features that do not exist at lower levels emerge~\cite{lehn2002toward}. For example, at the neurons level, single neuron is simple and its generalization capability is very poor. But when a certain number of such neurons are accumulated into a certain elaborate structure by certain ingenious combinations, the layers at the higher level begin to get the unprecedented ability of classification and feature learning. More importantly, such new gained capability or property is deducible from but not reducible to constituent neurons of lower levels. It's not a property of the simple superposition of all constituent neurons, and the whole is greater than the sum of the parts. In systems theory, such kind of phenomenon is known as \textit{whole emergence} \cite{wierzbicki2015systems}. Whole emergence often comes from the evolution of the system~\cite{arthur1993evolution}, in which a system develops from the lower level to the higher level, from simplicity to complexity. In this paper, the motivation of NOR structures is to introduce a new structure level to RNN-related networks by transferring traditional RNN from the system to the agent and from the outer dimension to the inner dimension \cite{fromm2004emergence}.

In 1993, W. Brian Arthur \cite{arthur1993evolution} has identified three mechanisms by which complexity tends to grow as systems evolve:
\begin{itemize}
	\item Mechanism 1: increase in \textit{co-evolutionary diversity}. The agent in the system seem to be a new instance of agent class, type or species. As a result, the system seems to have new external agent types or capabilities.
	\item Mechanism 2: increase in \textit{structural sophistication}. The individual system steadily accumulates increasing numbers of new systems or parts. Thus, newly formed system seems to have new internal subsystems or capabilities.
	\item Mechanism 3: increase by ``\textit{capturing software}''. The system capture simpler elements and learns to ``program'' these as ``software'' to be used as its own ends. 
\end{itemize}
In this paper, with the guidance of first two mechanisms, we introduce two methodologies to NOR structures design, which are named as \textit{aggregation} and \textit{specialization}. Aggregation and specialization are natural operations for increasing complexity in complex systems \cite{fromm2004emergence}. The former is related to Arthur's second mechanism, in which traditional RNNs are aggregated and accumulated into a high-level layer in accordance with a specific structure, and the latter is related to Arthur's first mechanism, in which the RNN agent in a high-level layer is specialized as the RNN agent that performs a specific function. 

We make several implementations and carry out experiments on three different tasks, including sentiment classification, question type classification and named entity recognition. Experimental results show that our models outperform constitute simple RNN remarkably with the same number of parameters and achieve even better results than GRU and LSTM sometimes.

\section{Background}

\subsubsection{Systems Theory}

Systems Theory was originally proposed by biologist Ludwig Von Bertalanffy \cite{von1968general,von1972history} for biological phenomena. In biology systems, there are several different levels which begin with the smallest units of life and reach to the largest and most extensive category: molecule, cell, tissue, organ, organ system, organization etc. Traditionally, a system could be decomposed into its individual components so that each component could be analyzed as an independent entity, and components could be added in a linear fashion to describe the totality of the system \cite{walonickgeneral}. However, Von Bertalanffy argued that we cannot fully comprehend a phenomenon by simply breaking it down into elementary parts and then reforming it. We instead need to apply a global and systematic perspective to underline its functionality \cite{mele2010brief}, because a system is characterized by the interactions of its components and the nonlinearity of those interactions \cite{walonickgeneral}. 

\subsubsection{Whole Emergence}

In systems theory, the phenomenon (i.e., the whole is irreducible to its parts) is known as \textit{emergence} or \textit{whole emergence} \cite{wierzbicki2015systems}. Emergence can be qualitatively described as: the whole is greater than the sum of the parts \cite{upton2014whole}. Or it can also be quantitatively expressed as: 
\begin{align}
&	W>\sum_{i}^{n}p_{i} \label{gst:equation:0}
\end{align}
where $ W $ is the whole of the system and consists of $ n $ parts, and $ \{p_{i}\}_{i=1 \cdots n} $ is the $ i $-th part. In 1972, Philip W. Anderson highlighted the idea of emergence in has article ``More is Different'' \cite{Anderson1972MoreID} in which he stated that a change of scale very often causes a qualitative change in the behavior of the system. For example, in human brains, when one examines a single neuron, there is nothing that suggests conscious. But a collection of millions of neurons is clearly able to produce wonderful consciousness. 

The mechanisms behind the emergence of complexity can be used to design neural network structures. One of the widely accepted reasons is the repeated application and combination of two complementary forces or operations: stretching and folding (in Physics term \cite{thompson2002nonlinear}), splitting and merging (in Computer Science term \cite{hannebauer2002autonomous}), or specialization and cooperation (in Sociology term). Merging or aggregating of agents means generally a number of (sub-)agents is aggregated or conglomerated into a single agent. Splitting or specializing means the agents are clearly separated from each other and each agent is constrained to a certain class or role \cite{fromm2004emergence}.

\subsubsection{Recurrent Neural Networks at the Edge of Chaos}

Recurrent Neural Networks (RNNs) \cite{Werbos1988GeneralizationOB,elman1990finding} are a class of deep neural networks that possess internal short-term memory due to recurrent feed-back connections between units, which makes them be able to process arbitrary sequences of inputs. Formally, given a sequence of vectors $ \{x_t\}_{t=1 \cdots T} $, the equation of Simple RNN \cite{elman1990finding} is:
\begin{align}
	&	h_t = f(Wx_t + Uh_{t-1}) \label{rnn:equation:0}
\end{align}
where  $ W $ and $ U $ are parameter matrices, and $ f $ denotes a nonlinearity function such as tanh or ReLU. For simplicity, the neuron biases are omitted from the equation.

Actually, RNNs can behave chaotically. There have been some works analysing RNNs theoretically or experimentally from the perspective of systems theory. \cite{sontag1997recurrent} provided an exposition of research regarding system-theoretic aspects of RNNs with sigmoid activation functions. \cite{bertschinger2004real} analyzed the \textit{computation at the edge of chaos} in RNNs and calculated the critical boundary in parameter space where the transition from ordered to chaotic dynamics takes place. \cite{Pascanu2013OnTD} employed a dynamical systems perspective to understand the exploding gradients and vanishing gradients problems in RNNs. In this paper, we obtain methodologies from systems theory to conduct structure designs of RNN related models.

\section{Network of Recurrent Neural Networks}
\label{sec:nor}

\subsection{Overall Architecture}

\begin{figure}[ht]
	\centering
	\includegraphics[width=\columnwidth]{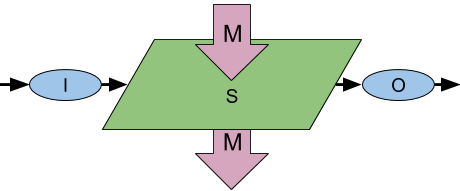}
	\caption{Overview of NOR structure.}
	\label{nor:illustration}
\end{figure}

\begin{figure*}[ht]
	\centering
	\subfigure[MA-NOR layer.]{
		\includegraphics[scale=0.9,width=0.22\linewidth]{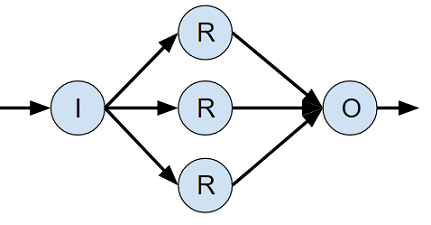}
		\label{aggragation:multi:agent}
	}
	\subfigure[Another MA-NOR layer.]{
		\includegraphics[width=0.24\linewidth]{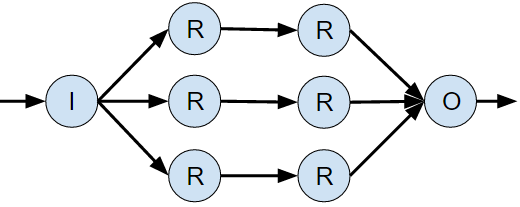}
		\label{aggragation:multi:agent:another}
	}
	\subfigure[MS-NOR layer.]{
		\includegraphics[width=0.24\linewidth]{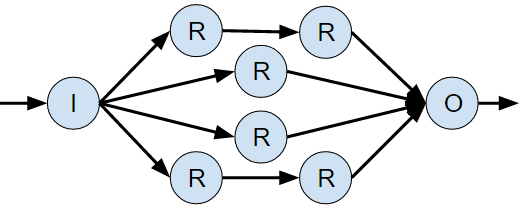}
		\label{aggragation:multi:scale}
	}
	\subfigure[SS-NOR layer.]{
		\includegraphics[width=0.24\linewidth]{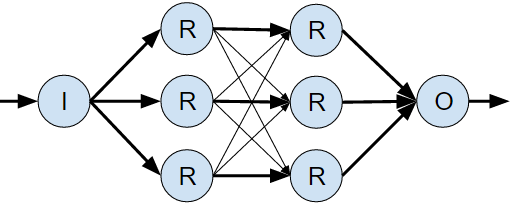}
		\label{aggragation:self:similarity:2}
	}
	\caption{The sectional views of NOR layers at one time-step. ``I'' means the component I, ``O'' means the component O, and ``R'' means RNN neuron.}
\end{figure*}

As the high-level illustration shown in Figure~\ref{nor:illustration}, NOR architecture is a three-dimensional spatial-temporal structure. We summarize NOR architecture as four components: I, M, S and O, in which component I (input) and O (output) control the head and tail of NOR layer, component S (subnetworks) is in charge of the spatial extension and component M (memories) is responsible for the temporal extension of the whole structure. We describe each component as follows:

\textbf{\textit{Component I}}: Component I controls the head of NOR architecture. It does data preprocessing tasks and distributes processed input data to subnetworks. At each time-step $ t $, the form of upcoming input data $x_t $ may be various, such as one single vector, or several vectors with the multi-granularity information, even the feature vectors with noise. One single vector may be the simplest situation, and the common solution is copying this vector into $ n $ duplicates and feed each of them into one single subnetwork in the component S. In this paper, the copying method meets our needs and we formalize it as:
\begin{align}
	&	\{x_{t}^{i}\}_{i=1 \cdots n} = C(x_{t}) \label{nor:comp:input}
\end{align}
in which $ C $ means copy function, and $ x_{t}^{i} $ will be fed into $ i $-th subnetwork.

\textbf{\textit{Component M}}: Component M manages all memories over the whole layer, not only internal but also external memories \cite{weston2014memory}. But in this paper, component M only considers internal memory and do not apply any extra processing to the individual memory of each RNN neuron. That is:
\begin{align}
	& m_{t}^{j} = I(o_{t-1}^{j}) \label{nor:comp:memory}
\end{align}
where $ I $ means identity function, the superscript $ j $ is the identifier of $j$-th RNN neuron \footnote{The notation of the ``subnetwork'' is different from the ``neuron'', for one subnetwork may be composed of several neurons. We use superscript $i$ as the identifier of the subnetwork. So, the input data of $i$-th subnetwork at time-step $t$ is denoted as $x_{t}^{i}$. }, $ m_{t}^{j} $ is the memory of $j$-th RNN neuron at time-step $ t $ and $ o_{t-1}^{j} $ is the transformation output of $j$-th RNN neuron at time-step $t-1$\footnote{In this paper, we just use simple RNN \cite{elman1990finding} applied with ReLU activation as our basic RNN neuron. Thus the memory at this time-step is just the output at last time-step.}.

\textbf{\textit{Component S}}: Component S is made up of several different or same subnetworks. Interaction may exist in these subnetworks. The responsibility of component S is to manage the logic of each subnetwork and handle the interaction between them. Suppose component S has $ n $ in-degree and $ m $ out-degree, i.e., component S receives $n$ inputs and produces $m$ outputs, the $ k $-th output is generated by necessary inputs and memories:
\begin{align}
	&	s_{t}^{k} = f([X, M]) \label{nor:comp:subnetowrk}
\end{align}
where $s_{t}^{k}$ is the $k$-th output at time-step $t$, $X$ and $M$ are needed inputs and memories, and $f$ is the nonlinear function which can be one-layer RNN or two-layer RNN etc.

\textbf{\textit{Component O}}: To form a layer we need a certain amount of neurons. So one of the NOR properties is multiple RNNs. A natural approach to integrate multiple RNN neurons' signals is collecting all outputs first and then using a MLP layer to measure the weights of each outputs. Traditional neuron outputs a single real value, so the collection method is directly arranging them into a vector. But RNN neurons is different, for each of them outputs a vector not a value. A simple method is concatenating all vectors and then connecting the concatenated vector to the next MLP. Another is pooling each RNN output vector into a real value, then arranging all these real values into a vector, which seems same as traditional neurons. In this paper, the former solution is used and formalized as:
\begin{align}
	&	 s_{t} = [ s_{t}^{1}; s_{t}^{2}; \cdots; s_{t}^{m} ] \label{nor:comp:output:1} \\
	&	 o_{t} = r(W_{MLP} * s_{t})  \label{nor:comp:output:2}
\end{align}
where $ s_t $ is the concatenated vector, $ W_{MLP} $ is the weight of MLP and $ r $ means the ReLU activation function of MLP.

\subsection{Methodology \uppercase\expandafter{\romannumeral 1}: Aggregation}

Any operation with changing a boundary can cause a emergence of complexity. The natural boundary is the agent itself, and sudden emergence of complexity is possible at this boundary if complexity is transfered from the agent to the system or vice versa from the system to the agent. There are two basic operations, aggregation and specialization, that can be used to transfer complexity between different dimensions \cite{fromm2004emergence}.

According to Arthur's second mechanism, internal complexity can be increased by aggregation and composition of sub-agents, which means a number of RNN agents is conglomerated into a single big system. In this way, aggregation and composition transfer traditional RNN from the outer to the inner dimension, from the system to the agent, for the selected RNNs are accumulated to become a part of a larger group. 

For a concrete NOR layer, suppose it is composed of $ n $ subnetworks, and $i$-th subnetwork is made up of $ k^i $ RNN neurons. Then, at the time-step $ t $, given the input $ x_t $, the operation flow is as follows:
\begin{enumerate}
	\item Component I: copy $ x_{t} $ into $ n $ duplications using equation (\ref{nor:comp:input}), then we get $ x_t^1, x_t^2, \cdots, x_t^n $.
	\item Component M: deliver the memory of each RNN neuron from the last time-step to the current time-step using equation (\ref{nor:comp:memory}), then we get memories $ m_t^{1,1}, \cdots, m_t^{1,k^1}$ in first subnetwork and memories
	$ m_t^{2,1}, \cdots, m_t^{2,k^2} $ in second subnetwork, etc.
	\item Component S: for each subnetwork $i$, take advantage of the input $x_t^i$ and memories $m_t^{i,1}, \cdots, m_t^{i,k^i}$ to get the nonlinear transformation output:
	\begin{align}
		& s_t^i = f([x_t^i, m_t^{i,1}, \cdots, m_t^{i,k^i}])
	\end{align}
	then, we get $ s_t^1, s_t^2, \cdots, s_t^n$.
	\item Component O: concatenate all outputs by equation (\ref{nor:comp:output:1}) and use a MLP function to determine how much signals in each subnetwork to flow through the component O by equation (\ref{nor:comp:output:2}).
\end{enumerate}

Obviously, the number, the type and the interaction of the aggregated RNNs determine the internal structure or inner complexity of the newly formed layer system. Thus, we propose three kinds of topologies of NOR aggregation method.

\subsubsection{Multi-Agent}

In systems theory, the natural description of complex system is the multi-agent system created by \textit{replication} and \textit{adaptation}. ``Replication'' means to copy and reproduce a new RNN agent, and ``adaptation'' means they are not totally same and some changes on weights or somewhere else by variation can increase the diversity of the system. 

\begin{figure}[ht]
	\centering
	\includegraphics[width=0.7\linewidth]{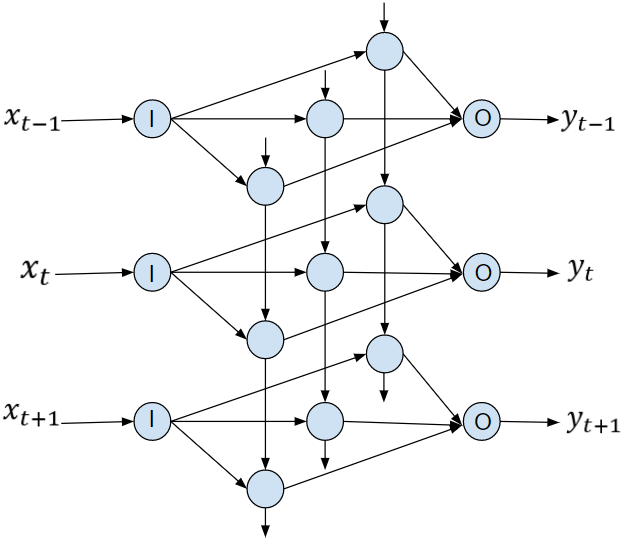}
	\caption{The unfolding of MA-NOR in three time-steps.}
	\label{aggragation:multi:agent:cubical}
\end{figure}

As shown in Figure \ref{aggragation:multi:agent}, there is a NOR layer (called MA-NOR) composed of four parallel RNNs. Figure \ref{aggragation:multi:agent:cubical} shows this layer being unrolled into a full network. Each subnetwork of MA-NOR layer is a one-tier RNN, thus at time-step $t$, the $i$-th subnetwork of component S in MA-NOR is calculated as:
\begin{align}
	s_t^i = o_t^i = r(W_i x_t^i + U_i m_t^{i})
\end{align}
where $r$ means ReLU activation function, $W_i$ and $U_i$ are parameters of corresponding RNN neuron, $o_t^i$ is the nonlinear transformation output and will be delivered to next time-step to be used as $m_{t+1}^{i}$, and $s_t^i$ is the output of $i$-th subnetwork which is equal to $o_t^i$.

The nonlinear function in equation (\ref{nor:comp:subnetowrk}) of each subnetwork may be more complex. For example, Figure \ref{aggragation:multi:agent:another} shows a NOR layer made up of three two-tier RNNs. At time-step $t$, the $i$-th subnetwork in component S is calculated as 
\begin{align}
	& o_t^{i,1} = r(W_t^{i,1}x_t^i + U_t^{i,1}m_t^{i,1})  \\
	s_t^{i} = & o_t^{i,2} = r(W_t^{i,2}x_t^i + U_t^{i,2}m_t^{i,2})
\end{align}

\subsubsection{Multi-Scale}

\begin{figure}[ht]
	\centering
	\includegraphics[width=0.7\linewidth]{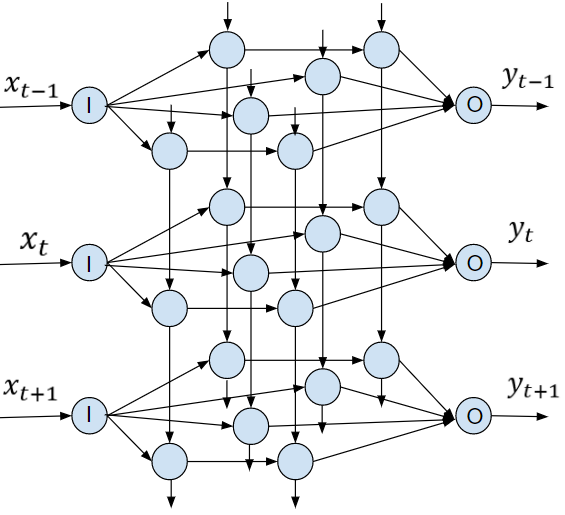}
	\caption{The unfolding of MS-NOR in three time-steps.}
	\label{aggragation:multi:scale:cubical}	
\end{figure}

The combination of multiple RNNs in a Multi-Agent NOR layer makes it somewhat like an ensemble. And empirically, diversity among the members of a group of agents is deemed to be a key issue in ensemble structure \cite{kuncheva2003measures}. One way to increase the diversity is to use the Multi-Scale topology which introduce new agent type to the system and can learn sequence dependencies in different timescales.

Figure \ref{aggragation:multi:scale} shows a NOR layer made up of four subnetworks, in which two of them are one-tier RNNs and the others are two-tier RNNs. Two kinds of timescale dependencies are learned in component S, which are formalized as follows:
\begin{align}
	s_t^{1} = & o_t^{1} = r(W_t^1 x_t^1 + U_t^1 m_t^{1}) \\
	s_t^{2} = & o_t^{2} = r(W_t^2 x_t^2 + U_t^2 m_t^2) \\
	& o_t^{3,1} = r(W_t^{3,1} x_t^3 + U_t^{3,1} m_t^{3,1}) \\
	s_t^{3} = & o_t^{3,2} = r(W_t^{3,2} o_t^{3,1} + U_t^{3,2} m_t^{3,2}) \\
	& o_t^{4,1} = r(W_t^{4,1} x_t^4 + U_t^{4,1} m_t^{4,1}) \\
	s_t^{4} = & o_t^{4,2} = r(W_t^{4,2} o_t^{4,1} + U_t^{4,2} m_t^{4,2})
\end{align}

\subsubsection{Self-Similarity}

The above mentioned aggregation and composition operation lead to big RNN agent-groups. While in turn, they can also be combined to form even bigger group. Such repeated aggregation and high accumulation makes the fractal and self-similar structure come into being.

\begin{figure}[ht]
	\centering
	\includegraphics[width=0.8\linewidth]{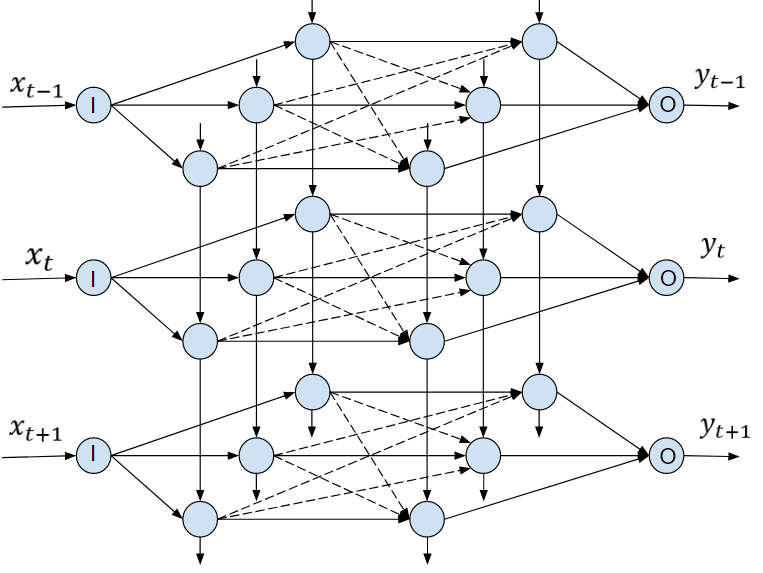}
	\caption{The unfolding of SS-NOR in three time-steps.}
	\label{ss:nor:cubical}
\end{figure}

As shown in Figure \ref{aggragation:self:similarity:2}, we also use three paths. But after each path first learns its own intermediate representation, the second layers gather all intermediate representations of three paths to learn high-level abstract features. In this way, different paths do not learn and train independently. The connections among each other helps the model easy to share informations. Thus, it becomes possible that the whole model learns and trains to be an organic rather than parallel independent structure. We formalize the cooperation of component S as follows:
\begin{align}
	& o_t^{1,1} = r(W_t^{1,1} x_t^1 + U_t^{1,1} m_t^{1,1}) \\
	& o_t^{2,1} = r(W_t^{2,1} x_t^2 + U_t^{2,1} m_t^{2,1}) \\
	& o_t^{3,1} = r(W_t^{3,1} x_t^3 + U_t^{3,1} m_t^{3,1}) \\
	s_t^{1} = & o_t^{1,2} = r(W_t^{1,2} [o_t^{1,1}, o_t^{2,1}, o_t^{3,1}] + U_t^{1,2} m_t^{1,2}) \\
	s_t^{2} = & o_t^{2,2} = r(W_t^{2,2} [o_t^{1,1}, o_t^{2,1}, o_t^{3,1}] + U_t^{2,2} m_t^{2,2}) \\
	s_t^{3} =&  o_t^{3,2} = r(W_t^{3,2} [o_t^{1,1}, o_t^{2,1}, o_t^{3,1}] + U_t^{3,2} m_t^{3,2}) 
\end{align}

\subsection{Methodology \uppercase\expandafter{\romannumeral 2}: Specialization}

We have mentioned that the emergence of complexity is usually connected to a transfer of complexity, a transfer at the boundary of the system. Aggregation and composition transfer complexity from the system to the agent, and from the outer dimension to the inner dimension. Another way to be used to cross the agent boundary is the specialization or inheritance, which transfer complexity from the agent to the system, and from the inner dimension to the outer dimension \cite{fromm2004emergence}. Specialization is related to Arthur's first mechanism. It increases structural sophistication outside of the agent by adding new agent forms. Through inheritance and specialization objects become objects of a certain class and agents become agents of a certain type, and the more such an agent becomes a particular class or type, the more it needs to delegate special tasks that it can not handle alone to other agents \cite{fromm2004emergence}. 

The effect of specialization is the emergence of delegation and division of labor in the newly formed groups. Thus, the formalization of $k$-th output in Component S can be rewritten as the following:
\begin{align}
& s_t^k = g(f_1([X, M]), f_2([X, M]), \cdots, f_L([X, M]))  \label{comp:spe:s}
\end{align}
where $f_l$ is the $l$-th specialized agent function, $ g$ means the cooperation of all specialized agents, and $L$ is the number of specialized agents. Equation (\ref{comp:spe:s}) denotes the function $f$ in equation (\ref{nor:comp:subnetowrk}) is implemented by the separated operations $f_1, f_2, \cdots, f_L$ and $g$.

\subsubsection{Gate-Specialization}

\begin{figure}
	\centering
	\includegraphics[width=1.\linewidth]{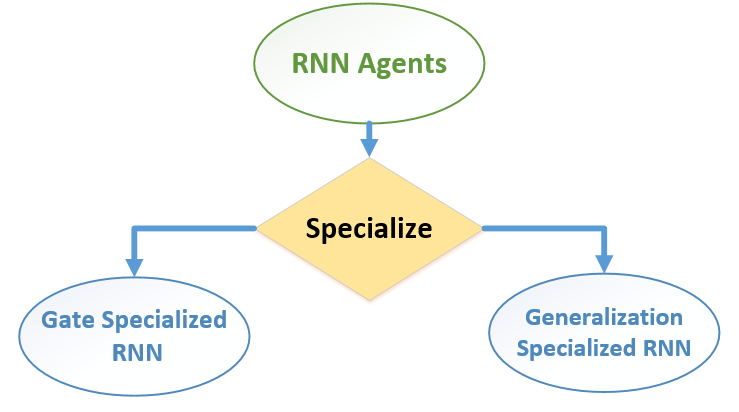}
	\caption{Gate Specialization.}
	\label{spe:exp}
\end{figure}

\begin{figure}[ht]
	\centering
	\includegraphics[width=\linewidth]{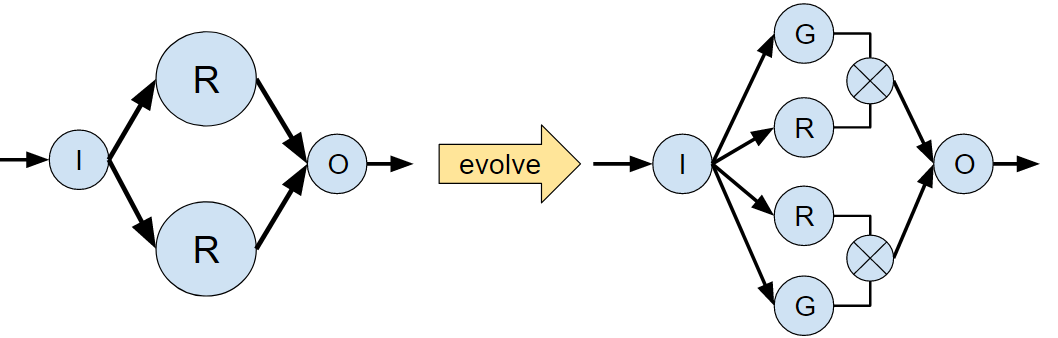}
	\caption{The sectional views of Gate-NOR layer at one time-step.}
	\label{gate:nor}
\end{figure}

We see gate mechanism is one of the specialization methods. As shown in Figure \ref{spe:exp}, a general RNN agent is separated into two specialized RNN agents, one is for gate duty and the other is for generalization duty. A concrete Gate-NOR is shown in Figure\ref{gate:nor}. In the original Multi-Agent NOR layer, each RNN agent is specialized as one generalization specific RNN and one gate specific RNN. We formalize it as:
\begin{align}
	& o_t^1 = \sigma(W_t^1 x_t^1 + U_t^1 m_t^1)\\
	& o_t^2 = r(W_t^2 x_t^2 + U_t^2 m_t^2)\\
	& s_t^1 = o_t^1 \odot o_t^2 \\
	& o_t^3 = \sigma(W_t^3 x_t^3 + U_t^3 m_t^3)\\
	& o_t^4 = r(W_t^4 x_t^4 + U_t^4 m_t^4)\\
	& s_t^2 = o_t^3 \odot o_t^4
\end{align}
where $ \sigma $ denotes the sigmoid activation and $ \odot $ denotes element-wise multiplication.

\begin{table*}[ht] 
	\footnotesize
	\centering
	\begin{tabular}{|c|c|c|c|c|c|c|c|c|}
		\hline
		
		\textbf{Task} & \textbf{\# of Params.} &
		\textbf{IRNN} & \textbf{GRU} & \textbf{LSTM} & \textbf{MA-NOR} & \textbf{MS-NOR} & \textbf{SS-NOR} & \textbf{Gate-NOR}\\ \hline
		
		\multirow{3}{*}{\textbf{Sentiment Classification}} 
		& 200 k & 212 & 107 &  88 &  89 & 66 & 61 & 61 \\
		& 400 k & 320 & 166 & 139 & 136 & 100 & 90 & 97 \\
		& 800 k & 468 & 252 & 213 & 203 & 149 & 132 & 149 \\ \hline
		
		\multirow{3}{*}{\textbf{Question classification}} 
		& 100 k & 198 & 86 & 68 & 74 & 54 & 53 & 45 \\
		& 200 k & 319 & 148 & 119 & 122 & 88 & 83 & 79 \\
		& 400 k & 497 & 244 & 199 & 193 & 139 & 126 & 133 \\ \hline
		
		\multirow{3}{*}{\textbf{Named Entity Recognition}}
		& 200 k & 197 & 86 & 67 & 74 & 54 & 53 & 45 \\ 
		& 400 k & 319 & 148 & 119 & 122 & 88 & 83 & 79 \\
		& 800 k & 497 & 244 & 199 & 193 & 139 & 126 & 133 \\
		 \hline
		
	\end{tabular}
	\caption{Number of hidden neurons for RNN, GRU, LSTM MA-NOR, MS-NOR, SS-NOR and Gate-NOR for each network size specified in
		terms of the number of parameters (weights).}
	\label{hidden:params}
\end{table*}

\subsubsection{Relationship with LSTM and GRU}

We see Long Short-term Memory (LSTM) \cite{hochreiter1997long} and Gated Recurrent Unit (GRU) \cite{chung2014empirical} as two special cases of Network of Recurrent neural networks. Take LSTM for example,  at time-step $ t $, given input $ x_{t} $ and previous memory cell $ c_{t-1} $ and hidden state $ h_{t-1} $, the transition equations of standard LSTM can be expressed as the following:
\begin{align}
&	i = \sigma(W_{i}x_{t} + U_{i}h_{t-1})  \label{lstm:equation:1} \\
& 	f = \sigma(W_{i}x_{t} + U_{f}h_{t-1}) \label{lstm:equation:2} \\
& 	o = \sigma(W_{o}x_{t} + U_{o}h_{t-1}) \label{lstm:equation:3} \\
&	g = tanh(W_{g}x_{t} + U_{g}h_{t-1}) \label{lstm:equation:4} \\
& 	c_{t} = c_{t-1} \odot f + g \odot i \label{lstm:equation:5} \\
& 	s_{t} = tanh(c_{t}) \odot o \label{lstm:equation:6}
\end{align} 
From the perspective of NOR (Network of Recurrent Neural Networks), LSTM is made up of four RNNs, in which three of four (i.e., $i$, $f$, $o$ RNNs) are specialized for gate tasks to control how much of informations let through in different parts. Moreover, there is only a shared memory $h_{t-1}$ which can be accessed by each RNN cell in LSTM.

While in turn, LSTM and GRU can also be combined to form even bigger group. 

\section{Experiments}
\label{sec:experiments}

In order to evaluate the performance of the presented model structures, we design experiments on the following tasks: sentiment classification, question type classification and named entity recognition. We compare all models under the comparable parameter numbers to validate the capacity of better utilizing the parametric space. In order to verify the effectiveness and universality of the experiments, we conduct three comparative tests under total parameters of different orders of magnitude, see Table \ref{hidden:params}. Every experiment is repeated 20 times with different random initializations and then we report the mean results. It's worthy noting that our aim here is to compare the model performance under the same hyper-parameter settings, not to achieve best performance for one single model.

\cite{le2015simple} showed that when initializing the recurrent weight matrix to be the identity matrix and biases to be zero, simple RNN composed of ReLU activation function (named as IRNN) can be comparable with even outperform LSTM. In our experiments, all basic RNN neurons are simple RNNs applied with ReLU function. We also keep the number of the hidden units same over all RNN neurons in a NOR model. Obviously, our baseline model is a single giant simple RNN \cite{elman1990finding} applied with ReLU activation. At the same time, two improved RNNs (GRU \cite{chung2014empirical} and LSTM \cite{hochreiter1997long}) have been widely and successfully used in NLP in recent years, so we also choose them as our baseline models.

The pre-trained 300-D Glove 6B vectors\footnote{https://nlp.stanford.edu/projects/glove/} and 300-D Google News vectors\footnote{https://code.google.com/archive/p/word2vec/} were obtained for the word embeddings. During training we fix all
word embeddings and learn only the other parameters in all models. The embeddings for out-of-vocabulary words are set to zero vectors. We pad or crop the input sentences to a fixed length. The trainings are done through stochastic gradient optimizer descent over shuffled mini-batches with the optimizer Adam \cite{kingma2014adam}. All models are regularized by using dropout \cite{srivastava2014dropout} method. At the same time, in order to avoid overfitting, early stopping is applied to prevent unnecessary computation when training. More details on hyper-parameters setting can be found in our codes, which are publicly available at $ \ddagger $.

\subsection{Sentiment Classification}
\label{experiment:sc}

We evaluate our models on the task of sentiment classification on the popular Stanford Sentiment Treebank (SST) benchmark~\cite{socher2013recursive}, which consists of 11855 movie reviews and is split into train (8544), dev (1101) and test (2210). SST provides detailed phrase-level annotation and all sentences along with the phrases are annotated with 5 labels: very positive, positive, neural, negative, and very negative. In our experiments, we only use the sentence-level annotation. One of our goals is to avoid expensive phrase-level annotation, like ~\cite{qian2016linguistically}. Another is, in practice, phrase-level annotation is hard to provide.

All models use the same architecture: embedding layer $ \rightarrow $ dropout layer $ \rightarrow $ RNN/NOR layer $ \rightarrow $ RNN/NOR layer $ \rightarrow $ max-pooling layer $ \rightarrow $ dropout Layer $ \rightarrow $ softmax layer. The first layer is the word embedding layer, next are two-layer RNN/NOR layers as the non-linear feature transformation layer. Then a max-pooling layer max-pools all transformed feature vectors by selecting the max value in each position to get sentence representation. Finally, a softmax layer is used as output layer to get the final result. To benefit from the regularization, two dropout layers with rate of 0.5 are added after embedding layer and before softmax layer. The initial learning rates of all models are set to 0.0002.  We use public available 300-D Glove 840B vectors to initialize word embeddings. Three different network sizes are tested for each architecture, such that the number of parameters are roughly 200 k, 400 k and 800 k (see Table \ref{hidden:params}). We set the minibatch size as 20. Finally, we use the Cross-Entropy criterion as loss function. 

\begin{table}[!htbp] 
	\footnotesize
	\centering
	\begin{tabular}{|c|c|c|c|}
		\hline
		\textbf{Model} & \textbf{200k Params} & \textbf{400k Params} &  \textbf{800k Params}\\ \hline
		\hline
		IRNN & 44.30 & 44.43 & 45.28 \\ \hline
		
		GRU & 47.27 & 47.35 & 47.65 \\ \hline 
		LSTM & 46.98 & 47.19 & 47.37 \\ \hline 
		\hline
		
		MA-NOR & 45.17 & 45.24 & 45.37 \\ \hline
		MS-NOR & 44.83 & 45.51 & 45.59 \\ \hline
		SS-NOR & 44.54 & 45.21 & 45.42 \\ \hline
		Gate-NOR & 44.95 & 45.80 & 46.06  \\ \hline
		
	\end{tabular}
	\caption{Accuracy (\%) comparison over different experiments on \textbf{SST} corpus.}
	\label{exp:sst:result}
\end{table}

The results of the experiments are shown in Table \ref{exp:sst:result}. It is obvious that NOR models get superior performances compared with IRNN baseline, especially when the network size is big enough. All models improve with network size grows. Among all NOR models, Gate-NOR gets the best results. However, we find that LSTM and GRU get much better results in three comparative tests.

\subsection{Question Type Classification}
\label{experiment:qtc}

Question classification is an important step in a question answering system which classifies a question into a specific type. For this task, we use TREC~\cite{li2002learning} benchmark, which divides all questions into 6 categories: location, human, entity, abbreviation, description and numeric. TERC provides 5452 labeled questions in the training set and 500 questions in the test. We randomly select 10\% of the training data as the validation set. 

\begin{table}[!htbp] 
	\footnotesize
	\centering
	\begin{tabular}{|c|c|c|c|}
		\hline
		\textbf{Model} & \textbf{100k Params} & \textbf{200k Params} &  \textbf{400k Params}\\ \hline \hline
		IRNN & 92.73 & 93.22 & 93.62 \\ \hline
		GRU & 91.83 & 92.54  &  93.64 \\ \hline
		LSTM & 91.44 &  92.46 & 93.10 \\ \hline \hline
		
		MA-NOR & 92.73 & 93.56 & 93.73 \\ \hline
		MS-NOR & 92.85 & 93.79 & 94.04 \\ \hline
		SS-NOR & 93.19 & 93.62 & 94.12  \\ \hline 
		Gate-NOR & 92.77 & 93.35 & 93.81 \\ \hline
		
	\end{tabular}
	\caption{Accuracy (\%) comparison over different experiments on \textbf{TREC} corpus.}
	\label{exp:trec}
\end{table}

All network types use the same architecture: embedding layer $ \rightarrow $ dropout layer $ \rightarrow $ RNN/NOR layer $ \rightarrow $ max-pooling layer $ \rightarrow $ dropout layer $ \rightarrow $ softmax layer. Dropout rates are set to 0.5. Three hidden layer sizes are chosen such that the total number of parameters for the whole model is roughly 100 k, 200 k, 400 k, see Table \ref{hidden:params}. All networks use a learning rate of 0.0005 and are trained to minimize the Cross Entropy Error. 

Table \ref{exp:trec} shows the accuracy of the different networks on the question type classification task. Here again, NOR models get better results than baseline IRNN model. Among all NOR models, SS-NOR also gets the best result. In this dataset, we find the performances of LSTM and GRU are even not comparable with IRNN, which proves the validity of results in \cite{le2015simple}.

\subsection{Named Entity Recognition}

Named entity recognition (NER) is a classic NLP task which tries to identity the proper names of persons, organizations, locations, or other entities in the given text. We experiment on CoNLL-2003 dataset \cite{tjong2003introduction} which consists of 14987 sentences in the training set, 3466 sentences in the validation set and 3684 sentences in the test set.

\begin{table}[!htbp] 
	\footnotesize
	\centering
	\begin{tabular}{|c|c|c|c|}
		\hline
		\textbf{Model} & \textbf{200k Params} & \textbf{400k Params} &  \textbf{800k Params}\\ \hline \hline
		IRNN & 85.07 & 85.52 & 85.58 \\ \hline
		GRU & 83.97 & 84.13 & 85.05 \\ \hline
		LSTM & 84.95 & 85.70 & 86.18 \\ \hline \hline
		
		MA-NOR & 85.96 & 86.17 & 86.18 \\ \hline
		MS-NOR & 85.51 & 86.10 & 86.21 \\ \hline
		SS-NOR & 86.06 & 86.20 & 86.33  \\ \hline 
		Gate-NOR & 85.67 & 85.89 & 86.21 \\ \hline
		
	\end{tabular}
	\caption{$ F_{1} $ (\%) comparison over different experiments on \textbf{CoNLL-2003} corpus.}
	\label{exp:ner}
\end{table}

Recently, popular NER models are based on bidirectional LSTM (Bi-LSTM) combined with conditional random fields (CRF), named as Bi-LSTM-CRF \cite{lample2016neural}. The Bi-LSTM-CRF networks can effectively use past and future features via a Bi-LSTM layer and sentence level tag information via a CRF layer. In our experiments, we also adapt this architecture by replacing LSTM with NORs or other variation of RNNs. So the universal architecture of all tested models is: embedding layer $ \rightarrow $ dropout layer $ \rightarrow $ Bi-RNN/Bi-NOR layer $ \rightarrow $ CRF layer. Three hidden layer sizes are chosen such that the total number of parameters for the whole network is roughly 200 k, 400 k and 800 k, see Table \ref{hidden:params}. We apply 50\% dropout after embedding layer. Initial learning rate is set to 0.005 and every epoch it is reduced by factor 0.95. The size of each minibatch is 20. We train all networks for 25 epochs and early stop the training when there are 5 epochs no improvement on validation set.

Our results are summarized in the Table \ref{exp:ner}. Not surprisingly, all NORs perform much better than giant single RNN-ReLU model. As we can see, GRU performs the worst, followed by IRNN. Compared to GRU and IRNN, LSTM performs very well, especially when network size grows up. At the same time, all NOR models get superior performances than IRNN, GRU and LSTM. Among them, SS-NOR model get best results.

\section{Conclusion}
\label{sec:conclusion}

In conclusion, we introduced a novel kind of systems theory based neural networks called ``Network Of Recurrent neural network'' (NOR) which views existing RNNs (for example, simple RNN, GRU, LSTM) as high-level neurons and then utilizes RNN neurons to design higher-level layers. Then we proposed several methodologies to design different NOR topologies according to the evolution of systems theory \cite{arthur1993evolution}. We conducted experiments on three kinds of tasks (including sentiment classification, question type classification and named entity recognition) to evaluate our proposed models. Experimental results demonstrated that NOR models get superior performances compared with single giant RNN models, and sometimes their performances even exceed GRU and LSTM.

\bibliographystyle{aaai}  
\bibliography{NetworkOfRecurrentNeuralNetworks} 

\end{document}